\documentclass[runningheads]{llncs}
\usepackage{graphicx}

\usepackage{graphicx}
\usepackage{marvosym}
\usepackage{caption}
\usepackage{subcaption}
\usepackage{multirow}
\usepackage{multicol}
\usepackage{amsmath}
\usepackage{xurl}
\usepackage{algpseudocode}
\usepackage{algorithm}
\usepackage[dvipsnames]{xcolor}
\usepackage{bm}

\begin{document}
\title{Robustness of AutoML on Dirty Categorical Data}
%
%
\author{Marcos L. P. Bueno\thanks{Affiliation by the time this work was done.} (\Letter) \and Joaquin Vanschoren }
\authorrunning{Bueno \& Vanschoren}
%
\institute{MCS, Eindhoven University of Technology, the Netherlands 
\email{marcoslbueno@gmail.com, j.vanschoren@tue.nl}}

\maketitle              
\begin{abstract}
The goal of  automated machine learning (AutoML) is to reduce trial and error when doing machine learning (ML). Although AutoML methods for classification are able to deal with data imperfections, such as outliers, multiple scales and missing data, their behavior is less known on dirty categorical datasets. These datasets often have several categorical features with high cardinality arising from issues such as lack of curation and automated collection.
Recent research has shown that ML models can benefit from morphological encoders for dirty categorical data, leading to  significantly superior predictive performance.
However the effects of using such encoders in AutoML methods are not known at the moment. 
In this paper, we propose a pipeline that transforms categorical data into numerical data so that an AutoML can handle categorical data transformed by more advanced encoding schemes.
We benchmark the current robustness of AutoML methods on a set of dirty datasets and compare it with the proposed pipeline. This allows us to get insight on differences in predictive performance. We also look at the ML pipelines built by AutoMLs  in order to gain insight beyond the best model as typically returned by these methods.

\end{abstract}
\newcommand{\aml}{AutoML}

\section{Introduction}
Automated machine learning methods (AutoML)  intend to reduce trial and error when using machine learning (ML) \cite{Hutter2019}. AutoML  techniques can select the best ML algorithm for the dataset at hand, tune its hyper-parameters, as well as  fix imperfections of the data, all automatically and in a data-driven way. The AutoML area has  seen several cases of success, e.g. in ML competitions \cite{Guyon2019} and in ML for health \cite{Waring2020}. AutoML pipelines (or simply pipelines, for brevity) can include preprocessing steps, e.g. for making the data suitable for a particular ML algorithm. For example, logistic regression classifiers require numerical and complete features, thus AutoML pipelines based on such  models include  operations to transform the data accordingly. Another situation is the inclusion of preprocessing  in pipelines for potentially improving the representation of the data, such as dimensionality reduction and outlier handling. 

Although quite some progress has been made in overall data preprocessing in AutoML methods, when it comes to handling categorical features the possible transformations are typically very simple. Standard approaches include e.g. one-hot encoding, ordinal encoding \cite{Hancock2020survey}, and less often (Bayesian) target encoding \cite{MicciBarreca2001}.
Research in ML indicates that real-life data with categorical features might have a multitude of specific problems, such as typos, lack of standards in acronyms and names, large number of distinct values (i.e. high cardinality), free-form fields, and so on \cite{Nazabal2020,cerda2018similarity,cerda2020encoding}. Such \textit{dirty} datasets are nowadays often encountered in real-life, e.g. when there is little to no data curation, automated data collection, and so on. Concrete examples include medical payments, traffic violations, and listings of unfinished drugs \cite{cerda2018similarity,cerda2020encoding}, where most of its features are categorical and have high-cardinality due to different causes. For training ML models on dirty data,  problems can arise when simple categorical encoders are used, such as substantial increase of sparsity, reduced predictive performance, hardware issues, among others.



More recently,  research has shown that specialized encoders for  dirty categorical data can provide substantial improvement to the performance of ML models \cite{cerda2018similarity,cerda2020encoding} in comparison to commonly used categorical encoders. These new encoders work by transforming categorical features into numerical features based on their morphological structure, e.g. by identifying similarities such as sub-words in categorical features.  However, little is known about how AutoML methods can benefit from specialized categorical encoders. One reason is that current AutoML benchmarks tend to focus on datasets with much less problematic categorical features \cite{zoller2021benchmark,Gijsbers2019bench}. Another important reason is that categorical encoders are more or less scalable mainly depending on the characteristics of the dataset at hand, which means that their placement in ML pipelines by AutoML must be taken carefully.
In order to achieve  a more sophisticated handling of categorical features in AutoML systems, we first need to understand the behavior of AutoMLs on dirty datasets, which is the central goal of this paper.


In this paper, we first evaluate the current robustness of AutoML methods on  dirty datasets, allowing us to understand if and when AutoMLs break down, and their predictive performance on such datasets.  Then, we introduce a categorical feature encoding that can leverage AutoML methods in datasets with categorical features,  and evaluate  how this affects their performance. We also look at the ML pipelines created by AutoMLs when using the proposed pipeline.  In this work, we use the GAMA AutoML \cite{Gijsbers2019}, which is an open-source AutoML based on genetic programming. 

The remainder of this paper is organized as follows. In Section \ref{sec:prelim} we briefly describe basic notions on categorical encoders and AutoML methods. Related work is discussed in Section \ref{sec:relwork}.  The proposed pipeline for categorical features and  its coupling to AutoMLs  are discussed in Section \ref{sec:methods}. The experimental setup and results on the current robustness of AutoMLs  as well as the proposed pipeline are provided in Section \ref{sec:results}.  The final remarks are given in Section \ref{sec:conc}.

\section{Preliminaries} \label{sec:prelim}
In this section we describe well-known encoding schemes (encoders, for brevity), i.e. procedures for transforming a categorical feature into a numerical feature. One possible taxonomy is as follows: an encoder is \textit{supervised} or \textit{target-based} if it takes the output feature into account during encoding, assuming that there is such a feature, e.g. a class feature; on the other hand, an encoder is \textit{unsupervised} or \textit{target-agnostic} if such information is not used to encode a categorical feature \cite{MicciBarreca2001}. However, we will look at encoders from a different perspective: whether the encoder tries to handle the \textit{cardinality} issue, e.g. by looking at the actual content of the categorical features.
To simplify the exposition, we denote by \textit{levels} or \textit{categories} the  set of distinct values of a categorical feature. 

\subsection{Simple categorical encoders}
We start off describing standard encoders that are  often used in AutoML methods. For other encoders, both supervised and unsupervised, we refer the reader to \cite{Hancock2020survey,McGinnis2018,Pargent2022regularized}.

\subsubsection{One-hot encoding}
The main idea of one-hot encoding, or \textit{dummy} encoding,  is that a categorical features with $n$ levels is transformed into $n$ new binary features, where the cells of the $i$-th new feature are set to 1 on the rows where the $i$-th categorical level  appeared originally, otherwise the cells are set to 0. One-hot encoding does not use class feature during encoding, hence it is an unsupervised method.

The dataset dimension can greatly increase when using  one-hot encoding, a problem that is often managed by combining  one-hot encoding with dimensionality reduction techniques  \cite{Hancock2020survey,MicciBarreca2001}. As an alternative,  the \textit{hash encoding} \cite{cerda2018similarity} can be used, which maps the original categories into a space of smaller dimension by means of a hash function.

\subsubsection{Ordinal encoding}
The ordinal encoding, also known as \textit{label encoding} \cite{Hancock2020survey}, transforms the categorical feature  by mapping its $n$ categories into integer numbers, typically  the set $\{0,\ldots,n-1\}$ \cite{sklearn}. The result is a single new feature that replaces the original categorical feature, hence no change in the dataset dimension occurs. No target feature is used during encoding. Despite its simplicity, the ordinal encoding scheme creates an artificial numerical order  in the new feature, since the original categories are typically unordered.

\subsection{Encoders for high-cardinality features}
We now describe  encoders that take into account characteristics of the levels of the categorical feature under consideration. The \textit{target encoding}  handles the cardinality issue by computing frequencies of levels, while the  \textit{dirty-cat encoders} look at the morphological structure of the levels, which includes \textit{similarity encoding}, \textit{min-hash encoding} and \textit{Gamma-Poisson encoding}.

\subsubsection{Target encoding}
The target encoding is based on the frequencies of categories on  the class feature \cite{MicciBarreca2001}. Assuming the class feature is binary, the posterior probability of a class given the category is computed. The posterior is often combined   with class priors computed over the entire training data, by summing a weighted  posterior with a weighted prior in  an Empirical Bayes manner \cite{MicciBarreca2001}. This is especially useful for high-cardinality features where the posterior probabilities might be unreliable due to low frequency of levels.   By replacing each category with its posterior probability possibly blended with the class prior,  a new numeric feature is obtained.  


Note that target encoding  does not change the dimensionality of the dataset, and is a supervised encoding method. Target encoding can also be extended to continuous or multi-class targets  \cite{MicciBarreca2001}. In this case, the target is first mapped onto one-hot encoding, then probabilities are computed. As a result, the dimension of the datasets is changed in the multi-class case.


\subsubsection{Similarity encoding}
The similarity encoding works by comparing  the levels of a categorical feature in terms of a string distance measure \cite{cerda2018similarity}. For example, the 3-grams of two levels are computed,  then a Jaccard similarity is used to indicate how similar these levels are. In the basic similarity encoding each level is compared to all the others, which leads to $n$ new features. This is why the similarity encoding can be seen as an  one-hot encoding based on  morphological information.  To reduce the dimensionality issue,  similarity encoding can be combined with  clustering and prototypes \cite{cerda2018similarity}.

\subsubsection{Min-hash encoding}
This scheme also looks at the string content of the categories, however a similarity measure for any two levels (e.g. the Jaccard similarity) is not computed directly. Instead, the similarity is approximated using a hashing function applied to the n-grams of the two levels. The minimum of these hash values is taken, and this process is repeated for a number of times as specified by the user, each time using a different hashing function. This procedure avoids comparing  n-grams directly, making the min-hash encoding more scalable.

\subsubsection{GAP encoding}
The Gamma-Poisson encoding (GAP) uses a topic model to learn  latent topics from the combination of sub-strings that occur frequently together in the data. The latent topics become  new features, and the encoded vectors correspond to the activations of the latent topics. Although the encoding produces numeric features, the GAP encoding is interpretable since the new features correspond to the latent topics, which are made of sub-strings of the original categories.

\subsection{Automated machine learning}
The goal of automated machine learning methods, or AutoMLs, is to reduce trial and error when doing machine learning. Since there is no single ML algorithm that works best for all datasets, we can formulate the problem of finding the best ML algorithm for a given dataset as a search problem. Moreover, many ML algorithms heavily rely on the tuning of its hyperparameters, and this dependence can be larger for different datasets. One way to put together all  these aspects is the Combined Algorithm Selection and Hyperparameter optimization problem (CASH), whose goal is to optimize for the following loss function \cite{feurer-neurips15a}:

\begin{equation} \label{eq:1}
    A^{*}, \bm{\lambda^{*}} \in 
    \underset{A^{(j)} \in \mathcal{A}, \bm{\lambda} \in \Lambda^{(j)} }{\arg\min} \frac{1}{K} \sum_{i=1}^{K} \mathcal{L}(A_{\bm{\lambda}}^{(j)}, D_{train}^{(i)}, D_{test}^{(i)})
\end{equation}

In equation \ref{eq:1}, the space of algorithms is denoted by $\mathcal{A} = \{A^{(i)}, \ldots, A^{(R)}\}$, where each algorithm $A^{(j)}$ has hyperparameters over the domain $\Lambda^{(j)}$. Assuming that the dataset is denoted by $D$ , the loss $\mathcal{L}(A_{\bm{\lambda}}^{(j)}, D_{train}^{(i)}, D_{test}^{(i)})$ represents the loss of algorithm $A^{(j)}$ with hyperparameters $\bm{\lambda}$ when trained  on $D_{train}^{(i)}$ and evaluated in $D_{test}^{(i)} = D - D_{train}^{(i)}$, where $i$ denotes the iteration in a $K$-fold cross-validation. While this formulation is based on  $k$-fold cross-validation, AutoMLs can be also configured to use instead a holdout setting, for example.

In real-life datasets, preprocessing is often necessary, e.g. to convert the categorical features into numerical features, or to handle missing values. To that end, the CASH problem can be adapted by adding the space of preprocessors into the space of algorithms $\mathcal{A}$. One can note that a potential solution to CASH corresponds to  a \textit{machine learning pipeline}.

There are several AutoML methods for different purposes, most of them for classification and regression tasks. They differ in how they solve the CASH problem, the backend  ML libraries they rely on, and the language programming. Some of the prominent AutoML methods are auto-sklearn \cite{feurer-neurips15a}, based on Bayesian optimization and meta-learning, with a backend on scikit-learn in python; TPOT \cite{Olson2016tpot} and GAMA \cite{Gijsbers2019}, based on genetic programming, using scikit-learn as well; H2O \cite{LedellH2O} which is based on random search, and uses their own backend in Java.



\section{Related work} \label{sec:relwork}
This work revolves around the intersection of several topics, including dirty datasets, categorical data, and AutoML. We discuss next relevant papers and their relation to this work. 

Dealing with dirty datasets in principled ways has been attracting substantial efforts in machine learning research.
A recent survey on different problems encountered in modern dirty datasets can be found in \cite{Nazabal2020}, where a taxonomy of issues is proposed and case studies are discussed. Other problems of relevance include inference of feature types, e.g.\ ptype \cite{Ceritli2020ptype} and ptype-cat \cite{Ceritli2021ptype-cat}, and the identification of the correct way to split columns in tabular data \cite{vanderBurg2019}.

There is a growing body of research in the context of categorical data. Early encoders designed for high-cardinality features include, e.g., the target encoding \cite{MicciBarreca2001}. The morphological encoders (also known as dirty-cat encoders) that are used in our pipeline were originally proposed in \cite{cerda2018similarity,cerda2020encoding}. While target encoding is a supervised encoding method, the morphological encoders  are unsupervised methods. However, none of such encoders were  evaluated in the context of AutoML. A benchmark of categorical encoders for high-cardinality data is provided in \cite{Pargent2022regularized}, however it does not consider AutoML methods and does not evaluate the morphological encoders we use in this paper. 

AutoML methods have been extended for multiple situations, including concept drift \cite{Celik2021} and online learning \cite{Celik2022}, outliers \cite{Singh2022outlier}, imbalanced datasets \cite{Singh2022imbalanced} and multi-target problems \cite{Wever2021}. A broader usage and understanding of dirty categorical data in AutoML, however, is lacking. The encoding scheme for categorical features varies per AutoML method, and they tend to use simple encoders  not designed for high-cardinality features. For example, GAMA \cite{Gijsbers2019} uses ordinal and one-hot  encoding depending on the number of levels in the categorical feature, while auto-sklearn \cite{feurer-neurips15a} uses one-hot encoding. On the other hand, H2O \cite{LedellH2O} uses mainly one-hot with varied strategies to reduce dimensionality,  as well as ordinal encoding. One exception is AutoGluon \cite{Erickson2020Autogluon}, which distinguishes categorical from text features; for categorical ones, ordinal encoding is used, while for text features two options are possible: an n-gram scheme, or a transformer neural network model with pretrained NLP models. The n-gram option of AutoGluon is similar to the similarity encoding of \cite{cerda2018similarity}.


\section{Proposed pipeline} \label{sec:methods}
We propose in this section a pipeline that handles the categorical features of a dataset, in particular the features of high-cardinality. The pipeline encodes the categorical features  into numerical features, and the transformed dataset can be sent to an AutoML system, which  apply its own procedures, e.g.  feature engineering, outlier repair, hyperparameter tuning, etc. From a global perspective, when coupled to  the proposed pipeline, an AutoML becomes more robust to datasets with dirty categorical features.


One practical tool  the proposed pipeline relies on is the inference of feature type. This is because it needs to know which are the categorical features. Dirty datasets  often do not have annotations regarding  column types, and a manual inspection is infeasible when there are too many columns. Moreover, dirty features often have  mixture of types \cite{Nazabal2020,Ceritli2020ptype}, e.g. a float column with values \{2, 3, `10.8', `7.2'\}, or multiple representations for missing values such as NaN, NA, -1, 99, etc. These situations are not well-handled by standard type detection of data science libraries, such as python pandas. Because of this we include in the pipeline a procedure to infer feature types.

The proposed pipeline for categorical features is shown in Algorithm \ref{alga}. Inference of feature types is done in lines \ref{alga_infa}-\ref{alga_infb}. Next, the encoding of categorical features is performed according to an encoding scheme.  Once this is done, the new dataset $D'$ corresponds to the encoded categorical features, plus the non-categorical features, which remain unchanged. Typically, the number of rows  of $D'$ will not change, which is the case   for the encoders considered in this work, so  $N'=N$. On the other hand, in most cases, encoding a categorical feature will result in new features, so it will be often the case $M' \neq M$.


\begin{algorithm}
\caption{Pipeline for categorical features}
\label{alga}
\begin{algorithmic}[1]
\Statex \textbf{Input}: Dataset $D = \{\textbf{x}_i,y_i \}_{i=1}^N $, with  $N$ rows and $M$ features
\Statex \textbf{Output}: Dataset $D' = \{\textbf{x}_i,y_i \}_{i=1}^{N'} $, with  $N'$ rows and $M'$ features, whose categorical features were transformed
\Statex 
\State Load the dataset $D$
\State $\textbf{T} \gets \text{feature\_type\_inference}(D)$, \Comment{$T$ is a list of  length $M$}  \label{alga_infa}
\State $\textbf{T}_{cat} \gets \{t \in \textbf{T}: t \text{ is categorical}\}$
\State $\textbf{T}_{other} \gets \textbf{T} - \textbf{T}_{cat}$ \label{alga_infb}
\State Choose an encoding scheme for $\textbf{T}_{cat}$         \label{alga_a}
\For{each categorical feature $\textbf{x} = \{x_i\}_{i=1}^N$ } \Comment{$\textbf{x} \in \textbf{T}_{cat}$}
    \State Encode   $\textbf{x}$ with the chosen scheme \label{alga_b}
\EndFor 
\State Finally, let  $D'$ be the dataset with $N'$ rows and $M'$ features containing the transformed features, i.e.  $\textbf{T}_{cat}$ and the unchanged features, i.e. $\textbf{T}_{other}$

\end{algorithmic}
\end{algorithm}

\section{Experimental results} \label{sec:results}
In this section we describe preliminary results of the proposed pipeline for dirty categorical features.  We also inspect the ML pipelines generated by AutoML to gain additional insight on models. In this paper, we focus on the GAMA AutoML \cite{Gijsbers2019} described in Section \ref{sec:prelim}.

\subsection{Setup and data}
We use a holdout evaluation, with a split of 75\% and 25\% for training set and validation set respectively. 
The AutoML tool only has access to the training data, which is normally  further split  internally for the AutoML's own evaluations (see equation \ref{eq:1}). 
Once model training is finished, GAMA returns the best model found, and this model is evaluated on the validation set. The \textit{metric} which the AutoML is optimized on is accuracy, and the \textit{time budget} is 1 hour per dataset.

The dirty datasets used  in experiments are listed in Table \ref{tab:datasets}. The inference of feature types was done using ptype \cite{Ceritli2020ptype}. All these datasets are classification tasks, and were also used in previous studies on categorical encoders  \cite{cerda2018similarity,cerda2020encoding}. In what follows, we will show results for two scenarios: (1) current robustness of AutoML, and (2) robustness of AutoML coupled to the proposed pipeline.

\begin{table}[htb]
    \centering
    \begin{tabular}{l|cccc}
\textbf{Dataset} &	\textbf{Classes} & 	\textbf{Instances} &	\textbf{Features} & 	\textbf{Cat.\ features} \\ \hline
\textbf{Kickstarter}	& 2 &	331k &	13 &	5 \\
\textbf{Openpayments} &	2 & 460k & 	178 & 	102 \\
\textbf{Roadsafety} &	3 &	1M &	66 &	2 \\
\textbf{Trafficviolations} &	4 &	1.7m &	42 &	36 \\
\textbf{Drugdirectory} &	7 &	120k &	20 &	14 \\
\textbf{Midwest} &	9 &	2k &	27 &	26 \\
\textbf{Metobjects} &	19 &	476k &	53 &	42 \\ \hline
    \end{tabular}
    \caption{Datasets used in the empirical evaluation of AutoML.}
    \label{tab:my_label}
    \label{tab:datasets}
\end{table}

\subsection{Current robustness of AutoML}

Table \ref{tab:gama_nohelp} shows the results of GAMA on dirty datasets. We first note that GAMA was able to fully process most of the dirty datasets on its own, since these datasets are given directly to GAMA as is.  GAMA achieves perfect predictive performance on the binary classification datasets, while the performance is good on the other datasets where it is able to finish. In some datasets, timeout errors by GAMA occurred  resulting in it not honoring the time budgets. Despite these errors, GAMA managed to finish normally in those datasets. 

\begin{table}[htb]
    \centering
    \begin{tabular}{l|ccccc}
\textbf{Dataset} &	\textbf{completed} &	\textbf{acc} &	\textbf{logloss} &	\textbf{best pipeline} &	\textbf{nb pipelines} \\ \hline
\textbf{Kickstarter} &	Yes* &	1.0 &	$\approx 0$ &	Imputation+RandomForest &	44 \\ 
\textbf{Openpayments} &	Yes* &	1.0 &	$\approx 0$ &	Imputation+GradBoosting &	21 \\
\textbf{Roadsafety} &	Yes &	0.60 &	0.47 &	Imputation+GradBoosting &	19 \\
\textbf{Trafficviolations} &	 \textcolor{red}{No} &	\textcolor{red}{-} &	\textcolor{red}{-} & \textcolor{red}{-} &	\textcolor{red}{0} \\	
\textbf{Drugdirectory} &	Yes &	0.98 &	0.06 &	Imputation+RandomForest &	63\\
\textbf{Midwest} &	Yes & 	0.50 &	1.46 &	Scaler+RandomForest	& 514 \\
\textbf{Metobjects} &	 \textcolor{red}{No} &	\textcolor{red}{-} &	\textcolor{red}{-} & \textcolor{red}{-} &	\textcolor{red}{0} \\	
\hline
    \end{tabular}
    \caption{Current robustness of GAMA, 1 hour budget. \textbf{Yes*}: finished in spite of time errors; \textbf{No}: no models produced. Metric to optimize is \textbf{accuracy}; \textbf{logloss} values shown for completeness.}
    \label{tab:gama_nohelp}
\end{table}

\subsection{Robustness using the proposed pipeline}
Although the proposed pipeline was designed to work with all categorical features in the data, in this paper we shall consider only the most predictive categorical feature. This gives a first impression of the potential of using morphological encoders together with AutoML methods. We follow the approach of \cite{cerda2018similarity}, where feature importances were calculated from a random forest model, from which the most predictive categorical feature was selected. As opposed to \cite{cerda2018similarity}, we leave the remaining categorical features untouched instead of converting them to one-hot, which leaves room for the AutoML to decide on those features. 

We use the function \textit{SuperVectorizer} from the dirty-cat library, which employs heuristics to select when to use each of the 3 morphological encoders. In summary, it uses similarity encoding when there are few levels in the categorical feature (e.g. fewer than 40), and switches to either min-hash or GAP encoding for larger number of categories.

The results of GAMA using the proposed pipeline is shown in Table \ref{tab:2}.  GAMA was now able to finish on all datasets and achieved better performance with the proposed pipeline. In particular, in Midwest where the accuracy increased from 0.50 to 0.68, and  Trafficviolations and Metobjects for which no pipelines where produced without the proposed pipeline (Table \ref{tab:gama_nohelp}). On the other hand, GAMA built fewer pipelines when using the pipeline, which occurs because the morphological encoders (SuperVectorizer) creates more features when encoding the selected categorical feature.

\begin{table}[htb]
    \centering
\begin{tabular}{l|cccc}
\textbf{Dataset} &	\textbf{acc} &	\textbf{logloss} &	\textbf{best pipeline} &	\textbf{nb pipelines} \\ \hline
\textbf{Kickstarter} &		1.0 & 	$\approx 0$ &	Imputation+LogReg &	32 \\
\textbf{Openpayments} &		1.0 &	$\approx 0$ &	Imputation+GradBoosting	& 19 \\
\textbf{Roadsafety} &		0.60 &	1.05 &	Imputation+BernoulliNB	& 18 \\
\textbf{Trafficviolations} &	0.78 &	0.51 &	Imputation+Scaler+BernoulliNB	& 8\\
\textbf{Drugdirectory} & 	0.99 &	0.04 &	Imputation+RandomForest	& 44 \\
\textbf{Midwest} & 	0.68 & 0.97 &	Imputation+RandomForest &	304 \\
\textbf{Metobjects} &		0.66 &	1.74 &	Imputation+Scaler1+Scaler2+BernoulliNB	& 19 \\ 
\hline
    \end{tabular}
    \caption{Robustness of GAMA using the proposed pipeline, 1 hour budget. GAMA produced models on all datasets. Metric to optimize is accuracy, logloss values shown for completeness.}
    \label{tab:2}
\end{table}

\subsection{Analysis of pipelines}
\begin{figure}[htb]
    \centering
    \includegraphics[scale=0.38]{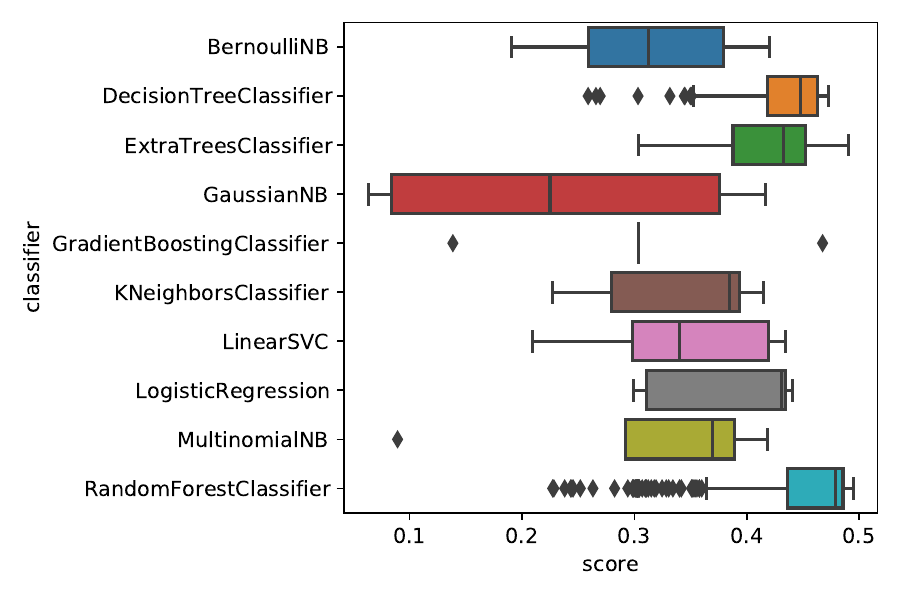}
    \includegraphics[scale=0.38]{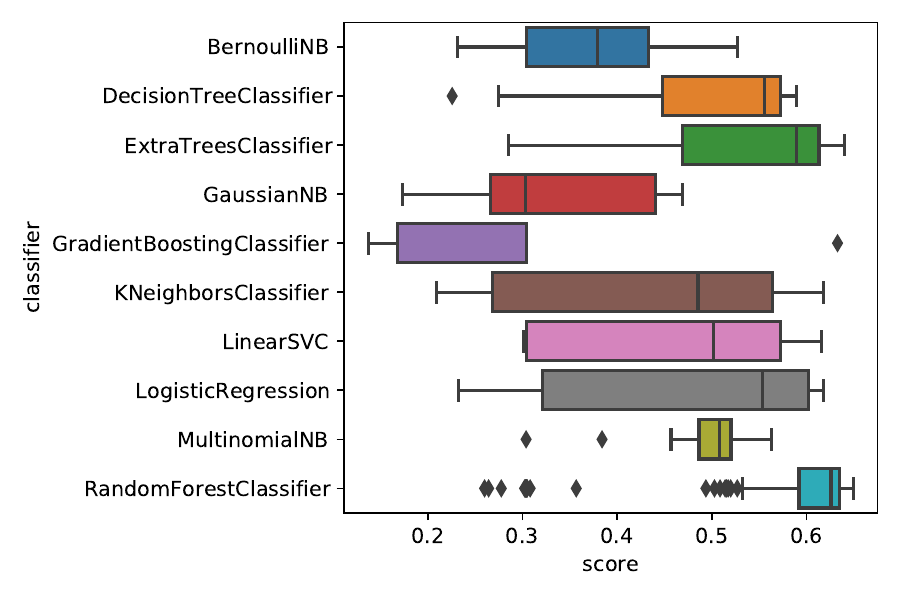}
    \caption{ML pipelines built by GAMA for the Midwest dataset. \textbf{Left}: original GAMA; \textbf{Right}: GAMA+new pipeline. Note the different scales.}
    \label{fig:current_pipelines}
\end{figure}

A summary of the performance of all pipelines built for the Midwest dataset is shown Figure \ref{fig:current_pipelines}. The variation in scores is caused by different choices of preprocessing and/or classifier hyper-parameters. The box plots give a loose idea  about the importance of tuning for this dataset. Besides the difference in performance, the two figures do not  indicate other substantial differences in terms of the performance of  different types of models for Midwest. One visual difference is that the variances become larger when using the proposed pipeline.


Such inspection of pipelines can also allow an AutoML user (e.g. a data scientist) to \textit{make choices}. One situation is when the user prefers pipelines with types of models different than those  in the best pipeline. In this case, the visual inspection can help the user select alternative pipelines while taking into account the  their (possibly reduced) predictive performance, and number and type of preprocessing used. For Midwest-left, the pipelines with tree-based models performed the best on this dataset: from all the 514 pipelines, the first 294 ones in the ranking are tree-based. The first non-tree model in the ranking is a logistic regression, with an accuracy of 0.44 (while the best overall achieved 0.50).


\section{Conclusion} \label{sec:conc}
In this paper we investigated the robustness of AutoML methods on dirty categorical data. We used the AutoML GAMA as a case study, and could assert that GAMA has a reasonable robustness based on experiments with a set of dirty datasets. We found that GAMA was able to process and produce models for most of the datasets, and achieved reasonable predictive performance. By using the proposed pipeline for categorical features, the predictive performance of GAMA was increased, and in some dirty datasets to a significant extent. These preliminary experiments used only the most predictive categorical feature, which shows great potential for further integration of morphological encoders into AutoML methods. This also leaves room for future evaluations on the AutoML performance if more categorical features were converted as well. For a broader benchmark, we also plan to use regression datasets, as well as extend the evaluation to other AutoML methods.

%
%
\bibliographystyle{splncs04}
\bibliography{biblio}

@article{Wever2021,
  author    = {Marcel Wever and
               Alexander Tornede and
               Felix Mohr and
               Eyke H{\"{u}}llermeier},
  title     = {AutoML for Multi-Label Classification: Overview and Empirical Evaluation},
  journal   = {{IEEE} Trans. Pattern Anal. Mach. Intell.},
  volume    = {43},
  number    = {9},
  pages     = {3037--3054},
  year      = {2021},
  doi       = {10.1109/TPAMI.2021.3051276},
  bibsource = {dblp computer science bibliography, https://dblp.org}
}

@inproceedings{Olson2016tpot,
  title={TPOT: A tree-based pipeline optimization tool for automating machine learning},
  author={Olson, Randal S and Moore, Jason H},
  booktitle={Workshop on automatic machine learning},
  pages={66--74},
  year={2016},
  organization={PMLR}
}

@misc{Erickson2020Autogluon,
  url = {https://arxiv.org/abs/2003.06505},
  author = {Erickson, Nick and Mueller, Jonas and Shirkov, Alexander and Zhang, Hang and Larroy, Pedro and Li, Mu and Smola, Alexander},
  title = {AutoGluon-Tabular: Robust and Accurate AutoML for Structured Data},
  publisher = {arXiv},
  year = {2020},
  copyright = {arXiv.org perpetual, non-exclusive license}
}

@article{LedellH2O,
    title = {{H2O} {A}uto{ML}: Scalable Automatic Machine Learning},
    author = {Erin LeDell and Sebastien Poirier},
    year = {2020},
    month = {July},
    journal = {7th ICML Workshop on Automated Machine Learning (AutoML)},
    url = {https://www.automl.org/wp-content/uploads/2020/07/AutoML_2020_paper_61.pdf},
}

@inproceedings{feurer-neurips15a,
    title     = {Efficient and Robust Automated Machine Learning},
    author    = {Feurer, Matthias and Klein, Aaron and Eggensperger, Katharina and Springenberg, Jost and Blum, Manuel and Hutter, Frank},
    booktitle = {Advances in Neural Information Processing Systems 28 (2015)},
    pages     = {2962--2970},
    year      = {2015}
}

@article{vanderBurg2019,
author = {van den Burg, G. J. J. and Naz\'{a}bal, A. and Sutton, C.},
title = {Wrangling Messy CSV Files by Detecting Row and Type Patterns},
year = {2019},
issue_date = {Nov 2019},
publisher = {Kluwer Academic Publishers},
address = {USA},
volume = {33},
number = {6},
issn = {1384-5810},
doi = {10.1007/s10618-019-00646-y},
journal = {Data Min. Knowl. Discov.},
month = {nov},
pages = {1799–1820},
numpages = {22}
}

@article{Singh2022imbalanced,
  author    = {Prabhant Singh and
               Joaquin Vanschoren},
  title     = {Automated Imbalanced Learning},
  journal   = {CoRR},
  volume    = {abs/2211.00376},
  year      = {2022},
  doi       = {10.48550/arXiv.2211.00376},
  eprinttype = {arXiv},
  eprint    = {2211.00376},
  bibsource = {dblp computer science bibliography, https://dblp.org}
}

@article{Singh2022outlier,
  author    = {Prabhant Singh and
               Joaquin Vanschoren},
  title     = {Meta-Learning for Unsupervised Outlier Detection with Optimal Transport},
  journal   = {CoRR},
  volume    = {abs/2211.00372},
  year      = {2022},
  doi       = {10.48550/arXiv.2211.00372},
  eprinttype = {arXiv},
  eprint    = {2211.00372},
  bibsource = {dblp computer science bibliography, https://dblp.org}
}

@article{Celik2022,
  author    = {Bilge Celik and
               Prabhant Singh and
               Joaquin Vanschoren},
  title     = {Online AutoML: An adaptive AutoML framework for online learning},
  journal   = {CoRR},
  volume    = {abs/2201.09750},
  year      = {2022},
  url       = {https://arxiv.org/abs/2201.09750},
  eprinttype = {arXiv},
  eprint    = {2201.09750},
  bibsource = {dblp computer science bibliography, https://dblp.org}
}

@article{Celik2021,
  author    = {Bilge Celik and
               Joaquin Vanschoren},
  title     = {Adaptation Strategies for Automated Machine Learning on Evolving Data},
  journal   = {{IEEE} Trans. Pattern Anal. Mach. Intell.},
  volume    = {43},
  number    = {9},
  pages     = {3067--3078},
  year      = {2021},
  doi       = {10.1109/TPAMI.2021.3062900},
  bibsource = {dblp computer science bibliography, https://dblp.org}
}

@inproceedings{Ceritli2021ptype-cat,
  title={ptype-cat: Inferring the Type and Values of Categorical Variables},
  author={Ceritli, Taha and Williams, Christopher K I},
  booktitle={21st ECML-PKDD Automating Data Science Workshop},
  year={2021},
}

@article{zoller2021benchmark,
  title={Benchmark and survey of automated machine learning frameworks},
  author={Z{\"o}ller, Marc-Andr{\'e} and Huber, Marco F},
  journal={Journal of artificial intelligence research},
  volume={70},
  pages={409--472},
  year={2021}
}

@inproceedings{Gijsbers2019bench,
  title={An open source AutoML benchmark},
  author={Gijsbers, Pieter and LeDell, Erin and Poirier, S{\'e}bastien and Thomas, Janek and Bischl, Bernd and Vanschoren, Joaquin},
  booktitle={6th ICML Workshop on Automated Machine Learning},
  year={2019}
}

@article{Ceritli2020ptype,
  title={ptype: probabilistic type inference},
  author={Ceritli, Taha and Williams, Christopher KI and Geddes, James},
  journal={Data Mining and Knowledge Discovery},
  volume={34},
  number={3},
  pages={870--904},
  year={2020},
  publisher={Springer}
}

@article{sklearn,
 title={Scikit-learn: Machine Learning in {P}ython},
 author={Pedregosa, F. and Varoquaux, G. and Gramfort, A. and Michel, V.
         and Thirion, B. and Grisel, O. and Blondel, M. and Prettenhofer, P.
         and Weiss, R. and Dubourg, V. and Vanderplas, J. and Passos, A. and
         Cournapeau, D. and Brucher, M. and Perrot, M. and Duchesnay, E.},
 journal={Journal of Machine Learning Research},
 volume={12},
 pages={2825--2830},
 year={2011}
}

@article{Pargent2022regularized,
  title={Regularized target encoding outperforms traditional methods in supervised machine learning with high cardinality features},
  author={Pargent, Florian and Pfisterer, Florian and Thomas, Janek and Bischl, Bernd},
  journal={Computational Statistics},
  pages={1--22},
  year={2022},
  publisher={Springer}
}

@article{Hancock2020survey,
  title={Survey on categorical data for neural networks},
  author={Hancock, John T and Khoshgoftaar, Taghi M},
  journal={Journal of Big Data},
  volume={7},
  number={1},
  pages={1--41},
  year={2020},
  publisher={SpringerOpen}
}

@article{McGinnis2018,
author = {McGinnis, William and Siu, Chapman and S, Andre and Huang, Hanyu},
year = {2018},
month = {01},
pages = {501},
title = {Category Encoders: a scikit-learn-contrib package of transformers for encoding categorical data},
volume = {3},
journal = {The Journal of Open Source Software},
doi = {10.21105/joss.00501}
}

@article{MicciBarreca2001,
author = {Micci-Barreca, Daniele},
title = {A Preprocessing Scheme for High-Cardinality Categorical Attributes in Classification and Prediction Problems},
year = {2001},
issue_date = {July 2001},
publisher = {Association for Computing Machinery},
address = {New York, NY, USA},
volume = {3},
number = {1},
issn = {1931-0145},
doi = {10.1145/507533.507538},
journal = {SIGKDD Explor. Newsl.},
month = {jul},
pages = {27–32},
numpages = {6},
}

@article{Gijsbers2019,
  doi = {10.21105/joss.01132},
  year  = {2019},
  month = {jan},
  publisher = {The Open Journal},
  volume = {4},
  number = {33},
  pages = {1132},
  author = {Pieter Gijsbers and Joaquin Vanschoren},
  title = {{GAMA}: Genetic Automated Machine learning Assistant},
  journal = {Journal of Open Source Software}
}

@article{Nazabal2020,
  title={Data engineering for data analytics: a classification of the issues, and case studies},
  author={Nazabal, Alfredo and Williams, Christopher KI and Colavizza, Giovanni and Smith, Camila Rangel and Williams, Angus},
  journal={arXiv preprint arXiv:2004.12929},
  year={2020}
}

@Inbook{Guyon2019,
author="Guyon, Isabelle
and Sun-Hosoya, Lisheng
and Boull{\'e}, Marc
and Escalante, Hugo Jair
and Escalera, Sergio
and Liu, Zhengying
and Jajetic, Damir
and Ray, Bisakha
and Saeed, Mehreen
and Sebag, Mich{\`e}le
and Statnikov, Alexander
and Tu, Wei-Wei
and Viegas, Evelyne",
editor="Hutter, Frank
and Kotthoff, Lars
and Vanschoren, Joaquin",
title="Analysis of the AutoML Challenge Series 2015--2018",
bookTitle="Automated Machine Learning: Methods, Systems, Challenges",
year="2019",
publisher="Springer International Publishing",
address="Cham",
pages="177--219",
isbn="978-3-030-05318-5",
doi="10.1007/978-3-030-05318-5\_10",
}

@book{Hutter2019,
  title={Automated machine learning: methods, systems, challenges},
  author={Hutter, Frank and Kotthoff, Lars and Vanschoren, Joaquin},
  year={2019},
  publisher={Springer Nature}
}

@article{Waring2020,
title = {Automated machine learning: Review of the state-of-the-art and opportunities for healthcare},
journal = {Artificial Intelligence in Medicine},
volume = {104},
pages = {101822},
year = {2020},
issn = {0933-3657},
doi = {https://doi.org/10.1016/j.artmed.2020.101822},
author = {Jonathan Waring and Charlotta Lindvall and Renato Umeton},
}

@article{cerda2018similarity,
  title={Similarity encoding for learning with dirty categorical variables},
  author={Cerda, Patricio and Varoquaux, Ga{\"e}l and K{\'e}gl, Bal{\'a}zs},
  journal={Machine Learning},
  volume={107},
  number={8},
  pages={1477--1494},
  year={2018},
  publisher={Springer}
}

@article{cerda2020encoding,
  title={Encoding high-cardinality string categorical variables},
  author={Cerda, Patricio and Varoquaux, Ga{\"e}l},
  journal={IEEE Transactions on Knowledge and Data Engineering},
  year={2020},
  publisher={IEEE}
}
\end{document}